\documentclass[10pt,oneside]{article}
\usepackage{amsfonts}
\usepackage{geometry}
\usepackage{graphicx}
\usepackage{textcmds}
\usepackage[ruled,vlined]{algorithm2e}
\usepackage{a4wide}

\usepackage[symbol]{footmisc}

\usepackage{listings}
\usepackage{xcolor}

\definecolor{codegreen}{rgb}{0,0.6,0}
\definecolor{codegray}{rgb}{0.5,0.5,0.5}
\definecolor{codepurple}{rgb}{0.58,0,0.82}
\definecolor{backcolour}{rgb}{0.95,0.95,0.92}

\lstdefinestyle{mystyle}{
    backgroundcolor=\color{backcolour},   
    commentstyle=\color{codegreen},
    keywordstyle=\color{magenta},
    numberstyle=\tiny\color{codegray},
    stringstyle=\color{codepurple},
    basicstyle=\ttfamily\footnotesize,
    breakatwhitespace=false,         
    breaklines=true,                 
    captionpos=b,                    
    keepspaces=true,                 
    numbers=left,                    
    numbersep=5pt,                  
    showspaces=false,                
    showstringspaces=false,
    showtabs=false,                  
    tabsize=2
}

\usepackage{subfigure}
\usepackage{url}
\usepackage{amsmath,amsthm,amsfonts,amscd, amssymb, epsf} 
\usepackage{geometry}
 \theoremstyle{plain}    
 \newtheorem{thm}{Theorem}[section]
 
 \numberwithin{equation}{section} 
 \theoremstyle{plain}
 \theoremstyle{plain}    
 \theoremstyle{definition}
 \newtheorem{defn}[thm]{Definition}
  
 \theoremstyle{plain}    
 \newtheorem{prop}[thm]{Proposition} 
  
 \newtheorem{lem}[thm]{Lemma}
 
 \newtheorem*{cor*}{Corollary}
 \newtheorem*{conj*}{Conjecture}
 \newtheorem*{thm*}{Theorem}

\newcommand{\bl}{\begin{lem}}
\newcommand{\el}{\end{lem}}

\newcommand{\bml}{\begin{multline}}
\newcommand{\eml}{\end{multline}}

\newcommand{\beq}{\begin{equation}}
\newcommand{\eeq}{\end{equation}}
\newcommand{\bp}{\begin{prop}}
\newcommand{\ep}{\end{prop}}
\newcommand{\bd}{\begin{defn}}
\newcommand{\ed}{\end{defn}}
\newcommand{\pf}{\begin{proof}}
\newcommand{\epf}{\end{proof}}

\title{Deep Learning Chromatic and Clique Numbers of Graphs}

\author{Jason Van Hulse and J.S. Friedman\footnote{The views expressed in this article are the author's  own and not those of the U.S. Merchant Marine Academy,the Maritime Administration, the Department of Transportation, or the United States government.}}

\begin{document}

\maketitle

\begin{abstract}

Deep neural networks have been applied to a wide range of problems across different application domains with great success. Recently, research into combinatorial optimization problems in particular has generated much interest in the machine learning community. In this work, we develop deep learning models to predict the chromatic number and maximum clique size of graphs, both of which represent classical NP-complete combinatorial optimization problems encountered in graph theory. The neural networks are trained using the most basic representation of the graph, the adjacency matrix, as opposed to undergoing complex domain-specific feature engineering. The experimental results show that deep neural networks, and in particular convolutional neural networks, obtain strong performance on this problem.
\end{abstract}

\section{Introduction}
Computing the chromatic number and clique number of a graph are classical NP-complete combinatorial optimization problems that have many applications to every day life. Recently, deep neural networks (DNN) have been utilized to develop heuristics in solving combinatorial optimization problems. DNNs offer several advantages as they naturally run in parallel on GPUs with many cores, and can be self-learning. Domain knowledge of the problem is not always necessary as the neural network can learn directly from supervised data and examples. 

In \cite{AMS19} the combinatorial optimization problem of solving Rubik's cube is studied. Here an A* search algorithm is implemented to unscramble a Rubik's cube that relies on a heuristic function to determine which node of the search tree to expand. The heuristic function is a DNN that is trained on a large set of examples, where each example results from scrambling the solved cube and recording how many turns it took to reach that scrambled state. The essence is that the neural network learns to look at a scrambled cube and determine how far it is from being solved. The A* search  utilizes this information to reach the solved cube goal. Two crucial points of \cite{AMS19} are that there is a unique goal state, and the problem is not NP-hard. Supervised examples can be generated efficiently to feed the neural network.

In \cite{scarselli2008} the Graph Neural Network (GNN) is introduced.  Nodes in a graph represent objects or concepts, and edges represent their relationships. Each concept is naturally defined by its features and the related concepts. A learning algorithm is given which estimates the parameters of the GNN model on a set of given training examples. In our paper we represent graphs as adjacency matrices and feed them to standard neural networks. Though the GNN is not as sensitive to node ordering as an adjacency matrix, our hope is that by feeding our neural networks enough training data, it can learn more invariant properties of graphs.

\cite{lemos2019} uses Graph Neural Networks (GNNs) to address the graph coloring problem. Their algorithm seeks to determine if a given graph admits a $\chi$-coloring, hence it is presented as a binary classification problem. The authors considered graphs with a vertex set size $N \in [40, 60]$, and chromatic number $\chi \in [3, 7]$. 

Inspired by the  {\em AlphaGoZero}  \cite{silver2017} algorithm,  \cite{huang19} uses a deep reinforcement learning algorithm for the graph coloring problem. By formulating the problem as a Markov Decision Process,   {\em AlphaGoZero} with graph embeddings is adapted to learn new heuristics in graph coloring. 

\cite{vesselinova2020learning} provides a survey of machine learning approaches across an array of combinatorial optimization problems on graphs such as maximum cut, graph coloring, and maximal clique. Both supervised and reinforcement learning methods are evaluated and compared.

In \cite{xing2020} a GNN is used to guide a Monte Carlo tree search in solving the traveling salesman problem (TSP) in the spirit of {\em AlphaGoZero}. Their results show shorter tours than other state of the art learning algorithms.

Our main goal is to use Deep Neural Networks (DNN) to approximate the chromatic number and clique number of a graph. Ultimately a goal would be to develop an algorithm analogous to \cite{AMS19} that can constructively color a graph and find the set of maximal cliques. However formidable obstacles make it difficult to directly apply \cite{AMS19}. Given a scrambled Rubik's cube, the minimal number of turns needed to unscramble the cube is a well defined function since there is a unique goal state to focus upon. In graph coloring there are many possible optimal colorings so it would be difficult to define a function that could measure how far a partially colored graph is from an optimal coloring. Another key difference is graph coloring is NP-complete, while the Rubik's cube is not. This makes it difficult to generate learning examples.

Let $G = (V,E)$ be a finite graph, where $V$ is the set of vertices, and $E$ is the set of edges; with \emph{order} $|V|,$ and \emph{size} $|E|.$  We consider vertex colorings of $G$ and let $\chi(G)$ denote the \emph{chromatic} number of $G,$  the minimal number of colors needed so that no two adjacent vertices are assigned the same color. Let $\omega(G)$ denote the \emph{clique} number, the maximum number of vertices in a maximum clique of $G.$ 

It is not clear as to the best way to input a graph to a neural network without losing its topological properties. Our hope was that the neural network would learn to see the edges connecting vertices, like a human would. As the adjacency matrices is square, and convolutional neural networks (CNN) are known to excel at computer vision, we felt they were a good place to start. We study various DNN architectures and measure how well they can learn  $\chi(G)$ or $\omega(G)$ for a graph $G.$ We generate a large number of examples of random graphs $G$ of order $N \leq 50$ with  $\chi(G)$ and $\omega(G)$ computed exactly. See \S\ref{secGenData}.  

After experimentation we were able to train CNNs to approximate the chromatic number and clique number with roughly 90\% accuracy (we defined success if we were off by one or less). See \S\ref{secResults}.

The complete code to generate our data and to implement the neural networks can be found at \url{github.com/mathprofessor/coloring1}.

\section{Generating Training Data} \label{secGenData}
\subsection{Generating Random Graphs}
Let $N \in \mathbb{N}.$ Denote the set of graphs of order $N$ by $X_N.$ When generating graphs of order $N$ we feel it is more natural for the DNN to also encounter graphs of all orders less than $N,$ so we actually generate graphs of various orders and embed them in $X_N.$ Let $n \in \{2,\dots,N\}.$ 

Let $K_n$ denote the \emph{complete} graph of order $n.$ Note that $V(K_n) = \{1,\dots,n \}$ and $|E(K_n)| = n(n-1)/2.$ Let $j$ be a uniformly random number between $0$ and $n(n-1)/2.$ To generate a random graph of order $n$ we shuffle the set of edges $E(K_n)$ and delete $j$ of the edges. Denote this graph by $G^0_n.$

Next append the vertices $\{n+1,\dots,N\}$ to $V(G^0_n)$ and do not modify the edges $E(G^0_n).$ Denote this new graph by $G^1_n.$ Note that it contains $N - n$ vertices that are isolated. It is of order $N$ but we wish to embed it in a more random manner into $X_N.$ Note that each of the vertices $\{n+1,\dots,N\}$ of $G^1_n$ has no edges incident with it.

Choose a random permutation $\sigma:\{1,\dots,N\} \mapsto \{1,\dots,N\}$ and define $E_n = \sigma(E(G^1_n)),$ where $\sigma(r,s) = \left(\sigma(r),\sigma(s)\right).$  Next define $G_n =  (\{1,\dots,N\},E_n).$

Finally for a given $k \in \mathbb{N},$ and $n \in \{2,\dots,N\}$ we generate $k$ different graphs $G_n.$ This will generate a training set with $k(N - 1)$ examples.  The Python code to generate our data sets is available at \url{github.com/mathprofessor/coloring1}. 

\subsection{Calculating the Chromatic Number and Clique Number}
Let $G$ be a graph. To compute $\omega(G)$ we utilize the algorithm of Bron and Kerbosch \cite{bron1973}. More specifically we encode our graph using the Python library NetworkX \cite{networkX} which implements an updated version \cite{bron1973,tomita2006,cazals2008}. 

Though computing maximal cliques and graph colorings are independent problems, we can utilize the maximal clique as a heuristic to speed up finding an optimal coloring using a SAT solver. Specifically we use the open source constraint programming solver Google OR-Tools: CP-SAT \cite{ortools}. 

Let $G = (V,E)$ and let $V = \{1,\dots,N\}.$ For $v \in V$ let $x_v$ be the color of vertex $v,$ where $x_v$ is integer valued with values also in $\{1,\dots,N\}.$ Let $z$ be an integer valued variable, also with values in $\{1,\dots,N\}.$ Suppose $G_c = (V_c,E_c)$ is a maximal clique of $G.$ 

We consider the following constraint programming problem with objective: 

\begin{enumerate}
\item For each $v$ in $V_c,$ arbitrarily assign to $x_v$  a unique color in $\{1,\dots,|V_c|\}.$
\item For each edge $(i,j) \in E,$ we add the constraint $x_i \neq x_j.$
\item For each vertex $i \in V,$ we have $x_i \leq z.$ 
\item Finally we add an objective to minimize the variable $z.$
\end{enumerate}

We summarize these steps in Algorithm~\ref{algGen}. The Python code to implement the above algorithm can be found at \url{github.com/mathprofessor/coloring1}.

\begin{algorithm}[H] \label{algGen}
\SetAlgoLined
\KwResult{Input $n,N.$ Creates random graph $G_n$ of size $n,$ randomly embeds in the space of graphs of size $N;$  calculates $\chi(G)$ and $\omega(G).$ }
Let $K_n$ be the complete graph of order $n;$

Set $G_0^n$ := The resulting graph after uniform randomly deleting $j$ edges of $K_n,$ where $0 \leq j \leq n(n-1)/2;$  

Set $G_1^n$ := the resulting graph after appending the vertices $\{n+1,\dots,N\}$ to the edges of $G_0^n;$

Choose a random permutation $\sigma$ of the vertices $\{1,\dots,N\}$ and use it to define an induced map on the edges of $G_1^n;$

Set $G:= G_n := \sigma(G_1^n);$

Set $G_c := (V_c,E_c)$ a maximal clique of $G;$

Color $(V_c,E_c)$ with colors $\{1,\dots,|V_c|\};$

Use a \emph{constraint program with objective} to optimally complete the coloring of $G.$
\\
 
 \caption{Generating random embedded graphs}
\end{algorithm}

\section{Exploratory Data Analysis}

Data was generated as described previously to create three disjoint datasets for training, validation and testing. The training set consists of 249,999 records, while the validation and test datasets each contain 24,999 records. 

\begin{figure}[h]
\begin{center}
    \subfigure{\includegraphics[width=0.48\textwidth]{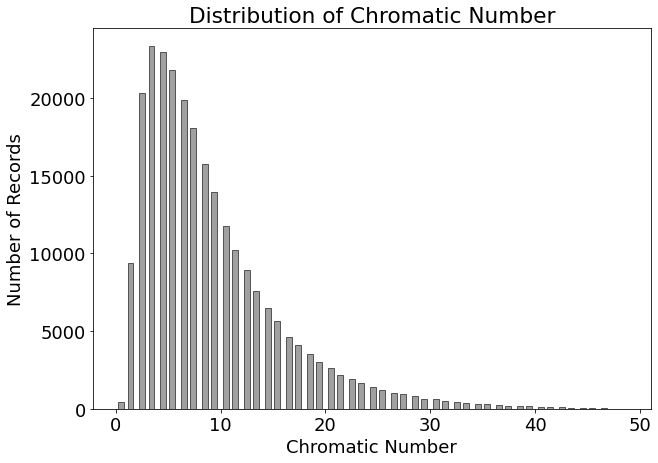} \label{fig:chromaticDist}}
    \subfigure{\includegraphics[width=0.48\textwidth]{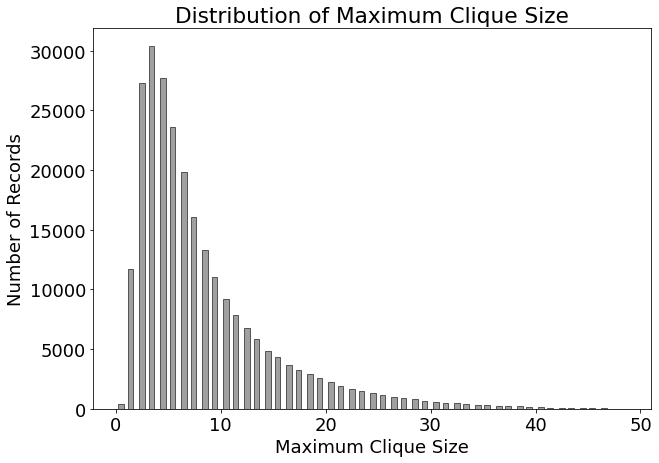} \label{fig:MaxCliqueDist}}
\end{center}\caption{Distribution of Chromatic Number and Maximum Clique Size}\label{fig:Dist}

\end{figure}

Figure~\ref{fig:Dist} shows the distribution of the chromatic number $\chi(G)$ and maximum clique size $\omega(G)$, respectively, for the training dataset. As can be seen from these graphs, the majority of graphs have $\chi(G)$ and $\omega(G)$ less than 10, though there is a long tail of larger values as well. The distributions for the validation and test datasets are very similar to what is presented in Figure~\ref{fig:Dist} and hence are omitted.

\section{Empirical Results}\label{secResults}

\subsection{Performance Metrics}

To measure and compare performance of the different models, the empirical study will use the {\em mean absolute error} (\textsc{mae}) and the percentage of records with an absolute error less than or equal to $l$, $P_l$ (typically, $l = 0.5 \mbox{ or } 1$):

$$
\textsc{mae} = \frac{1}{n} \sum \mid y_i - \hat{y}_i \mid
$$  

$$
P_l = \frac{\mid \{i : \mid y_i - \hat{y}_i \mid \le l \} \mid }{n}
$$

where $n$ is the number of records in the dataset and $y_i$ and $\hat{y}_i$ are the actual and predicted values of the dependent variable for record $i$. The absolute error for a single record $i$ is denoted as $\textsc{ae}(i)$. 

Training for the linear regression model uses only the training dataset.  Parameters for the deep neural network models are estimated using the training dataset, and the validation dataset is used to determine when to terminate the training process.  The test set is only used to estimate the generalization performance. \textsc{mae} is the metric used to train the neural network models, and $P_l$ is computed post-training for evaluation purposes.

\subsection{Linear Regression}

\begin{figure}[h]
\begin{center}
   \subfigure{\includegraphics[width=0.48\textwidth]{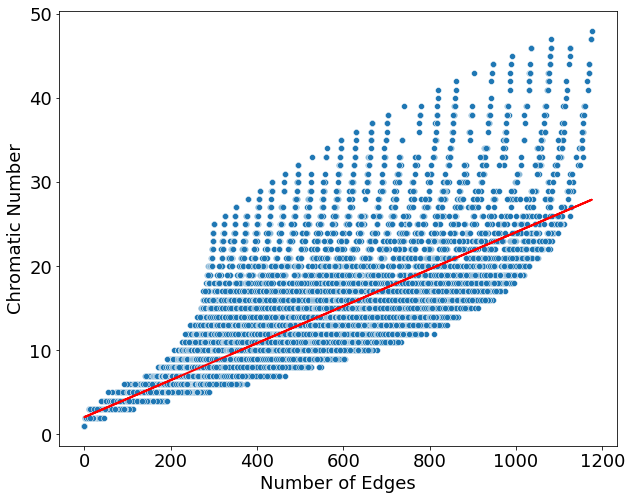} \label{fig:RegChromPred}}
    \subfigure{\includegraphics[width=0.48\textwidth]{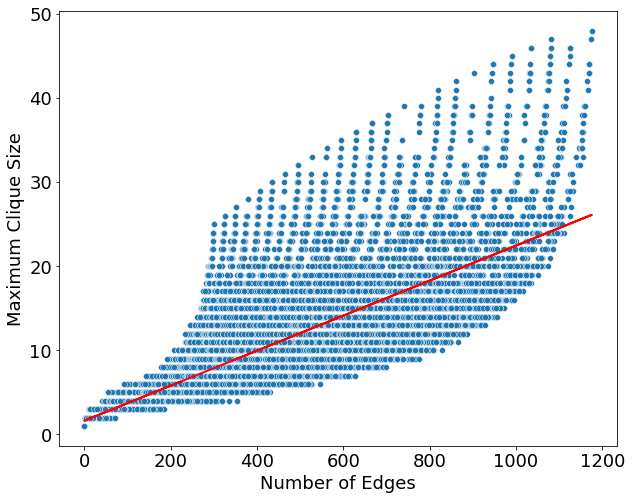} \label{fig:RegMCSPred}}
\end{center}\caption{Regression Model Baseline for predicting $\chi(G)$ and $\omega(G)$}\label{fig:Reg}
\end{figure}

To set a simple baseline for performance, we built a regression model with the independent variable as the number of edges in the graph, and the dependent variable either $\chi(G)$ or $\omega(G$). The scatterplots for the test data are in Figure~\ref{fig:Reg}, along with the regression line. 

Intuitively, there should be a positive correlation between the number of edges in a graph and both the chromatic number and maximum clique size. As can be seen from the scatterplot, for a fixed number of edges, there is a wide range of values for $\chi(G)$ and $\omega(G$).

\begin{table}[t]
\begin{tabular}{|c|c|c|}
\hline
Metric & $\chi(G)$ & $\omega(G)$ \\
\hline\hline
\textsc{mae} & 2.31 & 2.62 \\
\hline
$P_{0.5}$ & 26.3\% & 24.0\%\\
\hline
$P_1$ & 42.3\% & 40.1\%\\
\hline
\end{tabular}
\caption{Summarization of Regression Model Performance on Test Data}\label{tab:regPerf}
\end{table}

Table~\ref{tab:regPerf} summarizes the performance metrics of the regression model on test data. The mean absolute error on test data for predicting the chromatic number is 2.31, while the mean absolute error for predicting the maximum clique size is 2.62. There is certainly opportunity for substantial improvement over this very simplistic and naive model.


\subsection{Dense Deep Neural Network}

The objective of this work is to determine if a deep neural network can be built to predict $\chi(G)$ and $\omega(G$) without performing domain-specific feature engineering, hence using only the adjacency matrix of a graph. Our initial attempt is to use a dense network created using the {\em Sequential} API in Keras. For each input graph, the adjacency matrix was flattened into a $50 \times 50 = 2500$ input vector.  We experimented with various hyperparameters, including the number of hidden layers and the number of hidden nodes per layer, activation function, learning rate and optimizer. 

Generally speaking, lower values for the number of layers and number of hidden nodes per layer performed worse than higher values, while the choice of activation function, learning rate and optimizer did not substantially change the performance of the dense neural network. Therefore, our final dense model consisted of 13 hidden layers with 1000 nodes per layer, {\em ReLU} activation function, and {\em Adam} optimizer with the default learning rate. The final layer contains a single node with a linear activation function to produce a single output, either the chromatic number or maximum clique size. Note that all deep learning models were built in Tensorflow version~2.5 using the Keras API.

\begin{table}[t]
\begin{tabular}{|c|c|c|}
\hline
Metric & $\chi(G)$ & $\omega(G)$ \\
\hline\hline
\textsc{mae} & 1.59 &  1.81\\
\hline
$P_{0.5}$ & 37.5\% & 35.1\%\\
\hline
$P_1$ & 60.7\% & 57.7\%\\
\hline
\end{tabular}
\caption{Summarization of Dense Neural Network Model Performance on Test Data}\label{tab:dnnPerf}
\end{table}

The results for this dense model are presented in Table~\ref{tab:dnnPerf}. When predicting the chromatic number, the \textsc{mae} on the test dataset was 1.59, with $P_{0.5} = 37.5\%$ and $P_1 =60.7\% $.  Results for maximum clique size are similar: \textsc{mae} of 1.81, with $P_{0.5} = 35.1\%$ and $P_1 =57.7\%$. Compared to the results in Table~\ref{tab:regPerf}, this dense model presents significant lift over the regression model, as would be expected. 

\subsection{Convolutional Neural Networks}

Convolutional Neural Networks (CNNs) are commonly used with images, which are often represented as a 2-dimensional tensor of pixel values (in the case of greyscale images). In our case, the adjacency matrix of a graph is also 2-dimensional, with entries of 1 in the $i^{th}$ row and $j^{th}$ column if the two vertices $i$ and $j$ are connected, and 0 otherwise. Given this representation of the graph, CNNs are a natural choice for the architecture of the neural network. 

The input into the CNN is a tensor with dimensions $(50, 50, 1)$, representing the height (rows), width (columns), and number of channels.  We experimented with a number of different architectures, varying the number of convolutional and maximum pooling layers, the number of filters and size of the convolution kernel, in addition to the number of dense layers and hidden nodes per layer.  

Section~\ref{sec:BasicCNN} describes the Sequential CNN model that obtained solid performance on the test data, while Section~\ref{sec:WideCNN} describes a more refined architecture that we refer to as a {\em wide} convolutional neural network. 

\begin{table}[t]
\begin{tabular}{|c|c|c|}
\hline
Metric & $\chi(G)$ & $\omega(G)$\\
\hline\hline
\textsc{mae} &  0.43 &  0.55\\
\hline
$P_{0.5}$ & 68.0\% & 56.5\%\\
\hline
$P_1$ & 92.6\% & 81.6\%\\
\hline
\end{tabular}
\caption{Summarization of Sequential CNN Model Performance on Test Data}\label{tab:seqCNNPerf}
\end{table}

\subsubsection{Sequential Convolutional Neural Network}\label{sec:BasicCNN}

The architecture of the sequential CNN consists of a $Conv2D$ layer with a $3\times 3$ kernel and 512 filters, followed by a $MaxPool2D$ layer with $pool\_size = (2,2)$. A second $Conv2D$ layer with 64 filters and another $MaxPool2D$ layer follow. The output is flattened and fed into a dense network consisting of seven layers and 300 hidden nodes per layer. Finally, the activation function was set to $LeakyReLU$ for every layer in the network with the exception of the final dense layer, which consists of a single node with a linear activation function.

The performance of this model, trained separately for the chromatic number and maximum clique size, is shown in Table~\ref{tab:seqCNNPerf}. As can be seen from this table, the performance has improved substantially compared to the dense neural network: \textsc{mae} = $0.43$ for predicting $\chi(G)$ and $0.55$ for $\omega(G)$. 

\subsubsection{Wide Convolutional Neural Network}\label{sec:WideCNN}

\begin{figure}[h]
\begin{center}
\includegraphics[width=0.95\textwidth]{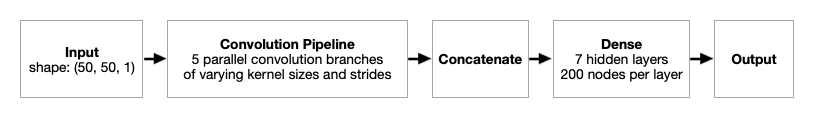} 
\end{center}\caption{High Level Architecture of the Wide Convolutional Neural Network} \label{fig:wideCNN}
\end{figure}

The high level architecture of the wide CNN model is shown in Figure~\ref{fig:wideCNN}. The second stage (after the input) consists of five parallel convolution pipelines with different size filters and strides, described in more detail below. The output of each of these five pipelines is flattened, and the results are concatenated together. The output of the concatenation layer is fed into a dense sequential network with seven hidden layers and 200 hidden nodes per layer, followed by a single dense layer with one node (and linear activation function) to create the output prediction. 

\begin{figure}[h]
\begin{center}
\includegraphics[width=0.95\textwidth]{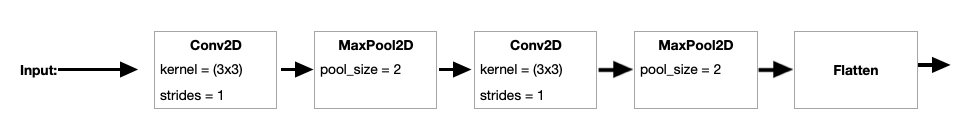} 
\end{center}\caption{First  convolution pipeline in the wide CNN}\label{fig:CNN_3_3_kernel}
\end{figure}

Figure~\ref{fig:CNN_3_3_kernel} shows the first pipeline of the convolution stage of the wide CNN model. The input data is processed through a $Conv2D$ layer with $kernel\_size = (3,3)$ and $strides = 1$, followed by a $MaxPool2D$ layer with $pool\_size = 2$ (note that we use $LeakyReLU$ activation functions throughout this network and hence omit from the diagram for simplicity). Additional $Conv2D$ and $MaxPool2D$ layers follow with the same parameters as before. The output is then flattened, and will be combined with the other convolution pipeline branches. 

\begin{figure}[h]
\begin{center}
\includegraphics[width=0.95\textwidth]{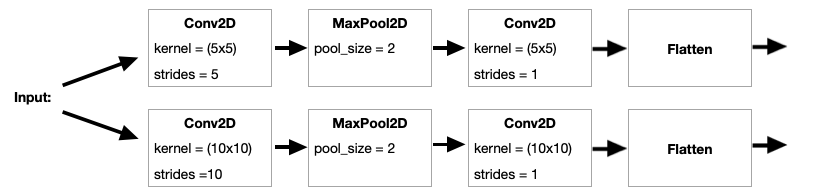} 
\end{center}\caption{Second and third convolution pipelines in the wide CNN}\label{fig:CNN_5_5_kernel}
\end{figure}

The second and third pipelines in the convolution stage are shown in Figure~\ref{fig:CNN_5_5_kernel}. The second branch uses a $5$ by $5$ kernel with $strides = 5$ for the first $Conv2D$ layer, followed by a $MaxPool2D$ layer. The second $Conv2D$ layer also uses a 5 by 5 kernel, but this time with a stride of 1. As before, the output of this $Conv2D$ layer is flattened. The third branch is similar to the second, except the kernel size is set to 10 by 10 with $strides = 10$ in the first $Conv2D$ layer. 

\begin{figure}[h]
\begin{center}
\includegraphics[width=0.6\textwidth]{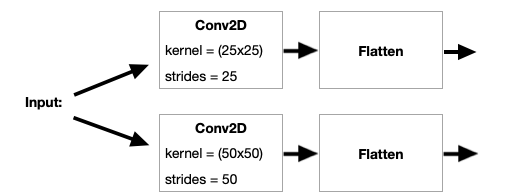} 
\end{center}\caption{Fourth and fifth convolution pipelines in the wide CNN}\label{fig:CNN_25_25_kernel}
\end{figure}

The fourth and fifth branches in the convolution pipeline are shows in Figure~\ref{fig:CNN_25_25_kernel}. Each $Conv2D$ layer uses very large kernels - 25 by 25 and 50 by 50, respectively. Finally, within all of the convolution pipelines, the first $Conv2D$ layer uses 512 filters, while the second uses $64$ filters. 

Compared to the sequential CNN in Section~\ref{sec:BasicCNN}, the wide CNN takes the foundational component of 2 convolutional layers with a kernel size of 3 by 3, fed into a dense network, and adds additional parallel convolutional layers with varying kernel sizes to the network. The idea is to use these varying kernel sizes to capture different relationships between vertices in the graph; indeed by considering only 3 by 3 kernels, relationships between larger subsets of vertices may be missed. 

Table~\ref{tab:wideCNNPerf} shows the results of the wide CNN model. When predicting the chromatic number, the model obtains an \textsc{mae} of 0.38 and $P_1$  of 90.9\%; in other words, over 90\% of the records in the test dataset had a prediction for the chromatic number within 1 of the true chromatic number. The \textsc{mae} of the wide CNN model for predicting the maximum clique size is 0.47, with $P_1 = 88.0\%$.

\begin{table}[t]
\begin{tabular}{|c|c|c|}
\hline
Metric & $\chi(G)$ & $\omega(G)$ \\
\hline\hline
\textsc{mae} &   0.38 &  0.47\\
\hline
$P_{0.5}$ & 68.7\% & 60.1\%\\
\hline
$P_1$ & 90.9\% & 88.0\%\\
\hline
\end{tabular}
\caption{Summarization of Wide CNN Model Performance on Test Data}\label{tab:wideCNNPerf}
\end{table}

\subsection{Discussion of Results} 

\begin{figure}[h]
\begin{center}
    \subfigure{\includegraphics[width=0.48\textwidth]{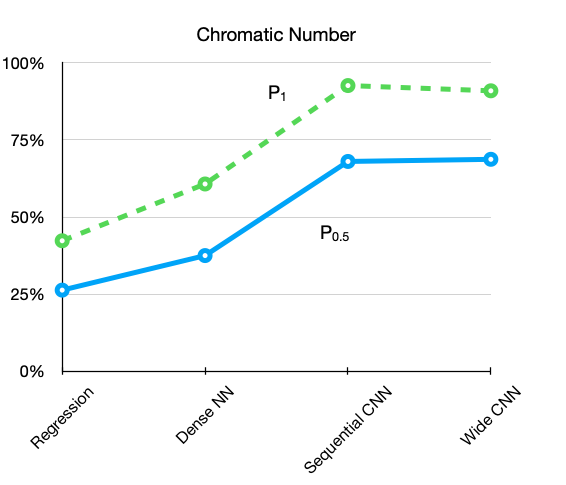} \label{fig:ChromNumSummary}}
       \subfigure{\includegraphics[width=0.48\textwidth]{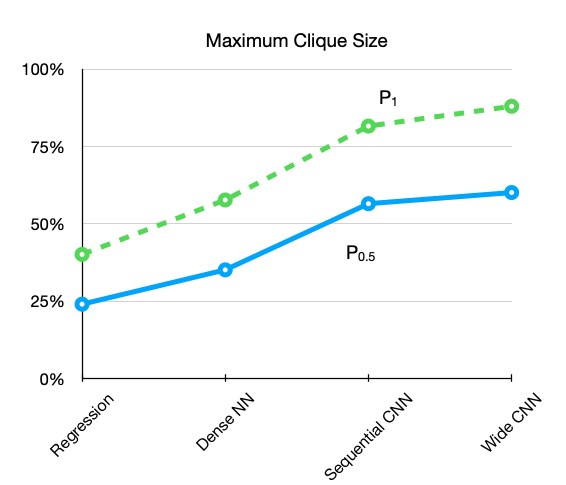} \label{fig:MCSSummary}}
\end{center}\caption{Summarized Comparison of Performance}\label{fig:SummaryComp}
\end{figure}

The performance of the four different models - a baseline regression model, a Dense Sequential Neural Network and two different Convolutional Neural Networks - is summarized in Figure~\ref{fig:SummaryComp} and in Table~\ref{tab:SummaryComp}.

Figure~\ref{fig:SummaryComp} shows the significant improvement in both metrics $P_1$ and $P_{0.5}$ towards the convolutional neural networks. Clearly the CNN architecture is detecting the relationships between nodes in the graph directly from the adjacency matrix in a way that is useful for predicting $\chi(G)$ and $\omega(G)$, showing substantial lift over a dense neural network.

\begin{table}[t]
\begin{tabular}{|l|cc|cc|}
\hline
 & \multicolumn{2}{|c|}{$\chi(G)$} & \multicolumn{2}{|c|}{$\omega(G)$} \\
\multicolumn{1}{|c|}{Model} & \textsc{mae} & \% Impr.  & \textsc{mae} & \% Impr. \\
\hline\hline
Regression &  2.31 & - &   2.62 & -\\
\hline
Dense NN & 1.59 & 31.2\% &1.81 & 30.9\%\\
\hline
Sequential CNN & 0.43 & 81.4\% & 0.55 & 79.0\%\\
\hline  
Wide CNN & 0.38 & 83.5\% & 0.47 & 82.1\%\\
\hline
\end{tabular}
\caption{Performance Comparison for the \textsc{mae} metric}\label{tab:SummaryComp}
\end{table}

Table~\ref{tab:SummaryComp} summarizes the \textsc{mae} values obtained on test data for each of the models. In addition, the column titled `\% Impr.' shows the percentage improvement in the mean absolute error versus the baseline (regression) model. For example, the Sequential CNN obtained an \textsc{mae} of 0.43, which is 81.4\% lower than the \textsc{mae} of the regression model. As with the data shown in Figure~\ref{fig:SummaryComp}, convolutional neural networks significantly outperform  other approaches. Further, the wide CNN, which combines convolution filters with different kernel sizes and strides, is able to combine information in a way that allows it to outperform a more standard CNN. 

\begin{figure}[h]
\begin{center}
\includegraphics[width=0.7\textwidth]{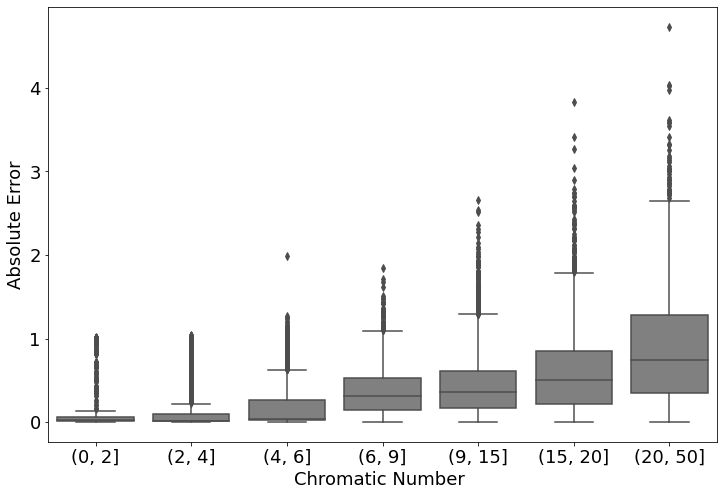} 
\end{center}\caption{Absolute Error Distribution by Chromatic Number}\label{fig:AbsErrorDistChrom}
\end{figure}

To provide a more in-depth understanding of the performance of the wide CNN, Figure~\ref{fig:AbsErrorDistChrom} shows the distribution of the absolute error for predicting $\chi(G)$ on the test data. To improve readability, records in the test data were first divided into groups by the value of $\chi(G)$ (e.g., $\chi(G) \in (0, 2]$).  A {\em boxplot} showing the distribution of absolute error for each group of records is then plotted. 

Each {\em box} shows three quartile values (e.g., $25^{th}$ percentile, median, and $75^{th}$ percentile) with whiskers that extend to 1.5 times the interquartile range (IQR).  As expected, the distribution of errors increases as the chromatic number increases, however performance of the model is very good regardless of the chromatic number.

To further analyze performance, an additional metric, the {\em absolute percentage error} (\textsc{ape}), is considered.  \textsc{ape} divides the absolute error by the actual target value, and is typically expressed as a percentage:

$$
\textsc{ape}(i) =  100 \times \frac{\mid y_i - \hat{y}_i \mid}{y_i }
$$  

Similar to \textsc{mae}, the \textsc{mape} is the mean absolute percentage error over all $n$ records in the test data:

$$
\textsc{mape} = \frac{1}{n} \sum \textsc{ape}(y_i)
$$  

Absolute percentage error is useful for counteracting the phenomenon observed in Figure~\ref{fig:AbsErrorDistChrom}, namely that as the value for the target variable increases, the observed absolute errors also tend to increase, but relative to the value of the target variable, the increases in error are not as meaningful. For example, if $y_i = 30$ and $\hat{y}_i = 32$, $\textsc{ae}(i)  = 2$ but $\textsc{ape}(i) = 6.7\%$. Compare this to a situation where $y_j = 2$ and $\hat{y}_j = 4$. Records $i$ and $j$ have the same absolute error ($\textsc{ae}(i) = \textsc{ae}(j) = 2$), however the absolute percentage error for $j$ is $100\%$, compared to $6.7\%$ for $i$.  We note that there are many additional performance metrics proposed in the literature (e.g., weighted \textsc{mape}, symmetric \textsc{mape}), however a full consideration of these metrics does not have a significant impact on this work.

\begin{table}[t]
\begin{tabular}{|l|c|c|}
\hline
\multicolumn{1}{|c|}{Model} & $\chi(G)$ & $\omega(G)$ \\
\hline\hline
Sequential CNN & 5.49\% & 7.41\%\\
\hline  
Wide CNN & 4.63\% & 6.69\%\\
\hline
\end{tabular}
\caption{Convolutional Neural Network performance measured by  \textsc{mape}}\label{tab:SummaryCompMAPE}
\end{table}

\begin{figure}[h]
\begin{center}
\includegraphics[width=0.7\textwidth]{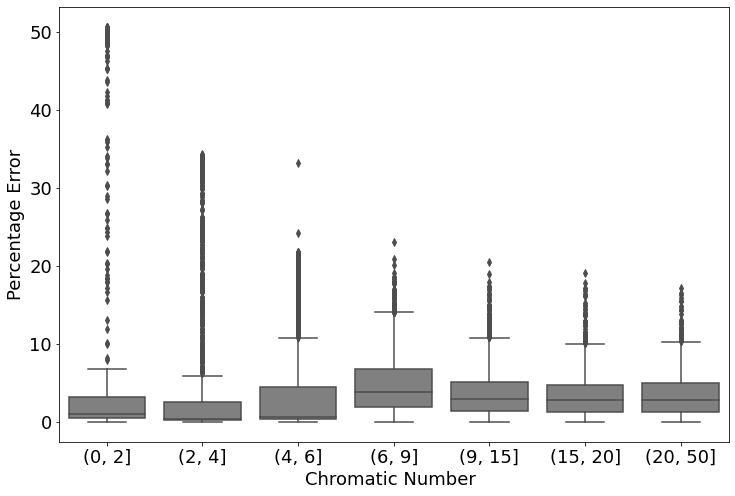} 
\end{center}\caption{Percent Error Distribution by Chromatic Number}\label{fig:PerErrorDistChrom}
\end{figure}

Table~\ref{tab:SummaryCompMAPE} shows the \textsc{mape} for both CNN models on the test data. All models achieve a \textsc{mape} significantly less than 10, which we would characterize as strong performance.  Furthermore, Figure~\ref{fig:PerErrorDistChrom} shows the distribution of \textsc{ape} obtained by the wide CNN across different ranges of $\chi(G)$, similar to Figure~\ref{fig:AbsErrorDistChrom}. As you can see from Figure~\ref{fig:PerErrorDistChrom}, the IQR is relatively stable across groups, but as expected, there is a higher level of skew in \textsc{ape} when $\chi(G)$ is small (i.e., when dividing by a small actual value, small absolute errors can become large percentage errors). 

\section{Conclusion}
Combinatorial optimization problems are of growing interest in the machine learning community, and in the context of graphs, graph coloring and clique search are two important areas of study. Searching for an optimal coloring and determining the maximum clique in a graph are known to be NP-complete, and research has begun to look at ways to improve heuristics to address these challenges.

In this research, we have trained deep learning models to directly predict the chromatic number and maximum clique size from the graph adjacency matrix. Since this is a supervised learning application, our training set consisted of graphs with known values for $\chi(G)$ and $\omega(G)$. The experimental results demonstrated conclusively that deep neural networks, and in particular convolutional neural networks, were able to learn from the training set and generalize well to previously unseen graphs of the same size. In particular, learning occurred without the need for domain-specific feature engineering of the graph input.

Future work will look to expand the scope of investigation to larger networks, where training data may be more limited due to the time involved in generating labeled data from large graphs. In addition, future work will consider the problem of graph coloring itself, i.e., coloring each node in a graph, as opposed to only predicting the number of colors needed.

\bibliography{fv1}
\bibliographystyle{amsalpha}

\vspace{5mm}

\noindent
Jason Van Hulse PhD \\
e-mail: jvanhulse@gmail.com \\

\vspace{5mm}
\noindent
Joshua S. Friedman PhD \\
Department of Mathematics and Science \\
\textsc{United States Merchant Marine Academy} \\
300 Steamboat Road \\
Kings Point, NY 11024 \\
U.S.A. \\
e-mail: friedmanJ@usmma.edu, crowneagle@gmail.com

\end{document}